\documentclass[fleqn,10pt]{wlscirep}
\usepackage[utf8]{inputenc}
\usepackage[T1]{fontenc}
\usepackage{multirow}
\usepackage{diagbox}
\usepackage{epstopdf}

\usepackage{amsmath}
\usepackage[normalem]{ulem}
\usepackage[table, xcdraw,dvipsnames]{xcolor}
\usepackage{soul}
\usepackage{textcomp}
\usepackage{siunitx}

\newcommand{\norm}[1]{\left\lVert#1\right\rVert}
\AppendGraphicsExtensions{.tif}

\title{An Autoencoder and Vision Transformer-based Interpretability Analysis of the Differences in Automated Staging of Second and Third Molars}

\author[1,*]{Barkin Buyukcakir}
\author[2]{Jannick De Tobel}
\author[3]{Patrick Thevissen}
\author[1]{Dirk Vandermeulen}
\author[1]{Peter Claes}

\affil[1]{KU Leuven, Department of Electrical Engineering (ESAT) - Processing Speech and Images (PSI), Leuven, 3000, Belgium}
\affil[2]{Ghent University, Department of Diagnostic Sciences, Ghent, 9000, Belgium}
\affil[3]{Imaging and Pathology - Forensic Odontology Department, KU Leuven, Leuven, 3000, Belgium}

\affil[*]{Corresponding author, barkin.buyukcakir@kuleuven.be}


\begin{abstract}
The practical adoption of deep learning in high-stakes forensic applications, such as dental age estimation, is often limited by the 'black box' nature of the models. This study introduces a framework designed to enhance both performance and transparency in this context. We use a notable performance disparity in the automated staging of mandibular second (tooth 37) and third (tooth 38) molars as a case study. The proposed framework, which combines a convolutional autoencoder (AE) with a Vision Transformer (ViT), improves classification accuracy for both teeth over a baseline ViT, increasing from 0.712 to 0.815 for tooth 37 and from 0.462 to 0.543 for tooth 38. Beyond improving performance, the framework provides multi-faceted diagnostic insights. Analysis of the AE's latent space metrics and image reconstructions indicates that the remaining performance gap is data-centric, suggesting high intra-class morphological variability in the tooth 38 dataset is a primary limiting factor. This work highlights the insufficiency of relying on a single mode of interpretability, such as attention maps, which can appear anatomically plausible yet fail to identify underlying data issues. By offering a methodology that both enhances accuracy and provides evidence for why a model may be uncertain, this framework serves as a more robust tool to support expert decision-making in forensic age estimation.
\end{abstract}
\begin{document}

\flushbottom
\maketitle
%
%
\newpage

\section*{Introduction}
The age of majority, or legal adulthood, is a critical threshold which affects criminal court decisions and punishment allocations, as individuals under this threshold receive significantly increased legal protection \cite{gummerum2020punishment}. Due to this importance, robust and reliable age estimation methods become highly desirable in forensic science, with many methods already defined in both the literature and in practice \cite{schmeling2016forensic}. In legal proceedings concerning living juveniles and young adults, where the age is highly impactful in the outcome, especially when proof of identity is lacking or the claimed age is suspected to be unreliable, dental age assessment is performed by forensic odontology experts \cite{matsuda2020forensic}.
The gold standard of age estimation in this age group is the assessment of dental development by allocating stages to permanent teeth. Due to well defined sequence of tooth development especially earlier in life, and in the case of the third molar around the age of majority \cite{lewis2015forensic}, dental age has proven to be a valuable proxy for chronological age. For forensic age estimation in living individuals, a panoramic radiograph is employed to capture the entire dentition and maxillomandibular complex, thereby also revealing potential developmental anomalies. Traditionally, the lower left permanent teeth are assessed, as well as all four third molars. Several manual staging schemes have been created, each with a varying number of stages and differences in stage membership criteria \cite{thevissen2013third,niu2021review,rahim2023reliability}. These manual techniques commonly are applied by having at least two experienced dentists assigning stage labels to teeth, and one additional expert acting as the tie breaker in the cases where a disagreement is encountered. As they rely heavily on expert knowledge, these techniques often suffer from intra- and inter-observer variability\cite{dhanjal2006reproducibility}. More specifically, the stage assessment for the same tooth can differ significantly between different observers and different evaluation sessions and agreement percentages as low as 64\% can be encountered \cite{pillai2021inter}. This variability motivates computer-aided, automated staging techniques that are, in most implementations, deterministic. 

In particular, deep learning models, which can learn complex relations between the inputs and the labels, have been the recent preferred method of automating this process. Convolutional neural networks (CNNs), with their intrinsic suitability for computer vision tasks, have been successfully applied in the broader field of medical image analysis in many studies \cite{anwar2018medical}. Specifically in dental stage allocation, De Tobel et al. (2017)\cite{de2017automated} used a pretrained AlexNet to estimate the stage of tooth 38 with 0.51 mean accuracy. Banar et al. (2020)\cite{banar2020towards} achieved an accuracy of 0.54 in the stage allocation of third molars, using a CNN-based segmentation and classification framework. Han et al. (2022)\cite{han2022or} reported a total accuracy of 0.87 while predicting the developmental stages of all right mandibular teeth using ResNet-101. A recent study by Matthijs et al. (2024)\cite{matthijs2024artificial} evaluated the DenseNet-201 architecture in the automated staging of all permanent tooth types, and reported an accuracy of 0.71 for tooth 37, and 0.57 for tooth 38. These studies underline the predictive capabilities of deep learning models in dental stage allocation.

While the usage of CNNs largely eliminates the problems related to manual staging, their practical use remains limited because of the large disparities in accuracy. Moreover, since CNNs are considered "black boxes" in the sense that they cannot explicitly express their decisions in a domain-related context,  \cite{sartor2020impact,kroll2015accountable,danks2017regulating,buyukccakir2024opg}, most studies have not provided an explanation for the low or high accuracy of automated staging. With the recent global focus by organisations such as the U.S. Defense Advanced Research Projects Agency (DARPA) \cite{national2019national}, and the European Union with its General Data Protection Regulation (GDPR) \cite{sartor2020impact}, on interpretable deep learning systems intensifying, explainability in deep learning became more important. 

Explainability in visual media is most commonly expressed by the generation of saliency/attention maps \cite{niu2021review}. These attention maps depict the varying degrees of impact the regions of the input image have on the decision, thus confirming or declining their agreement with human interpretation. 
These attention maps can be generated by investigating the gradient within a deep model w.r.t. the input image. Well-known methods with this approach to attention maps are guided backpropagation \cite{springenberg2014striving}, and the more recent Grad-CAM \cite{selvaraju2017grad}. However, there are several issues with the gradient-based approaches, such as their post-hoc nature, and the assumption that gradients reflect importance \cite{yasin2024is,woerl2023initialization}. A recent departure from such gradient-dependent attention maps has been the introduction of the self-attention mechanism of transformers \cite{vaswani2017attention}. Initially proposed for text-based learning tasks, the workflow of the transformer architecture relies on this attention mechanism to learn global spatial relations between text elements. This approach was adapted to visual tasks, which benefit from spatial attention in many cases, culminating in the proposal of the Vision Transformer (ViT) architecture \cite{dosovitskiy2020image}. This architecture reframes images as a sequence of patches. It uses linear embeddings augmented with positional encoding to feed information to a transformer encoder model, where the self-attention mechanism is in effect. This pipeline is then appended with a multi-layer perceptron head for downstream tasks, such as image classification. The inherent advantage of transformers in explainability is that the learned attention directly utilised in the learning task can be extracted, therefore not requiring additional steps after training to compute attention maps such as in gradient-based methods \cite{dosovitskiy2020image} while also achieving performance on par with CNNs \cite{mauricio2023comparing}.

Attention maps help explore the behaviour of deep models, and are the go-to method to establish visual explainability, to reveal the model decision process, and assess whether the basis for decisions is correct. An example, shown by Ribeiro et al. (2016)\cite{ribeiro2016should}, is when a model learned to focus on the snowy background to predict the label "wolf", without focusing on the animal in the images, showing the localisation of attention on non-relevant regions and indicating the model not to be trustable in practical application. However, equally frequently, the opposite attention failure mode can be observed where the attention map is "correct", but the model predictions are not \cite{ray2021generating}. This had led to the distinction of plausible explanations and faithful explanations \cite{jain2019attention,wiegreffe2019attention}. In general purpose applications such as ImageNet object classification, the object classes are quite different from each other, therefore a coarser attention map is usually sufficient. In applications such as medical image classification, however, the inter-class variations can be much more nuanced. In such applications, when attention maps only loosely correspond to expert references, even when accompanied by correct predictions, the confidence in the model diminishes \cite{saporta2021deep}, indicating the plausibility of the explanations do not guarantee faithfulness. Given these points, it can be seen in high-stakes applications, such as medical and forensic, where inter-class variation is small, it is diligent to employ attention maps along with supporting methods of interpretability \cite{rudin2019stop}.

As stated by Rudin (2019)\cite{rudin2019stop}, creating interpretable models, or at least not relying on black-box explanations, is the natural solution to this conundrum. Based on the preceding discussion of visual attention methods, we can state that these methods rely on the black box learning scheme for explanations and do not necessarily cover the general behaviour of the model itself.  We take a figurative step back and adopt this philosophy not only on a sample-by-sample explanation level but, more generally, to analyse model suitability, seeking to interpret why a model does or does not perform well. 

In this paper, we apply automated dental staging with a ViT model as classifier on two datasets of molar images, tooth numbers 37 and 38 (following the FDI World Dental Federation numbering scheme), individually. As reported by Matthijs et al. (2024)\cite{matthijs2024artificial}, there was a large performance disparity between teeth 37 and 38, even though they are both lower molars. In the current study we aim to verify this performance difference, and understand the reasons behind it by making use of several conventional DL methods. Ultimately, we aim to confirm the hypothesis that the performance difference in the automated staging of these two teeth is not due to the inability of the deep learning methods employed, but are due to data-related factors, or more specifically, due to the variable morphology of tooth 38\cite{al2023root,morita2016exploring}, hindering the predictive performance of the trained models. We thus propose a deep learning framework that can provide meta-information which human experts can use to confirm or deny that the model in question has learned to perform stage classification in line with current expert understanding. A secondary goal is to utilise the proposed framework in order to increase classification accuracy for both tooth 37 and tooth 38. We propose to supplement the attention map-based explanations with the latent space of the autoencoder model, and with visually inspectable image reconstructions. As a first step, we fine-tune and evaluate a pretrained DenseNet-201 model\cite{huang2017densely}, which is a popular CNN architecture, in order to establish a performance baseline, and to reproduce the results presented by Matthijs et al. (2024). Then, we train a ViT model on the images in order to analyze the self-attention values, visualized as attention maps. Finally, we train and evaluate our proposed framework, consisting of an autoencoder model trained with triplets, and a ViT, which is trained using the reconstructed images from the autoencoder model. We perform metric analysis on the latent space representations of teeth images, and the resulting attention maps in order to show that the model for tooth 37 learns to perform the staging task more reliably than tooth 38, and discuss the reasons for the low classification performance for the latter in light of this analysis.

\section*{Methods}

\begin{figure}[!htb]
    \centering
    \includegraphics[width=\textwidth]{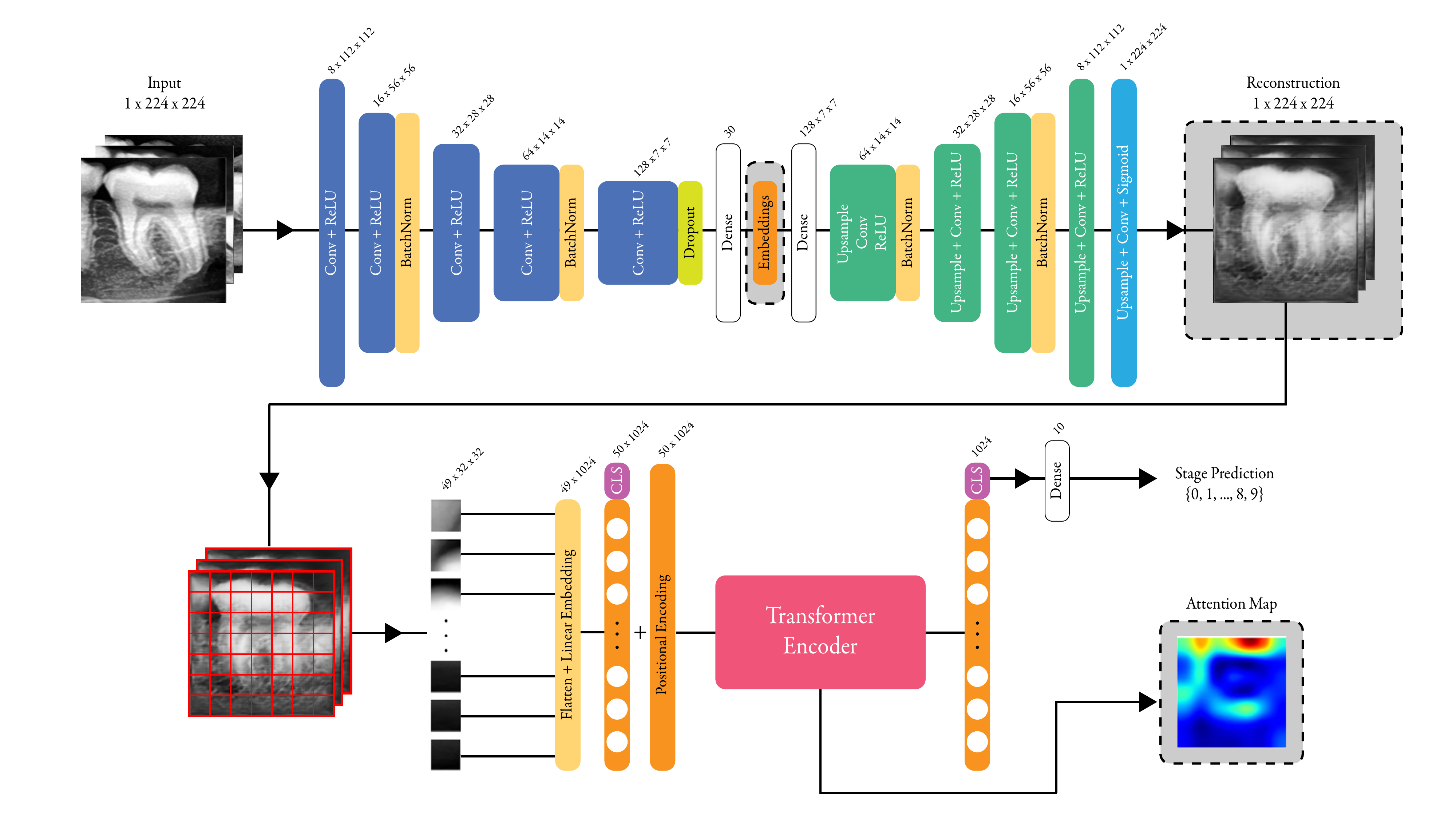}
    \caption{Overview of the complete pipeline. There are two sequential trainings taking place, the first being of the autoencoder (top) and the second of the ViT (bottom) on the image reconstructions. The autoencoder consists of a symmetric encoder-decoder setup of three layers each, with linear projections to and from the embedding space. The final layer of the decoder ends with a sigmoid activation in order to limit the pixel intensities between 0 and 1. The interpretability elements are marked with gray boxes with dashed borders. These elements (latent space embeddings, reconstructed images and the attention maps) help increase the transparency in model behaviour by providing bases for further analysis.}
    \label{fig:overview}
\end{figure}

Supplementary Table 1 depicts the Checklist for Artificial Intelligence in Medical Imaging (CLAIM) compliance of this work. Two datasets were evaluated, those of teeth number 37, a second molar and 38, a third molar. The datasets are exactly the same as the ones used by Matthijs et al. (2024). These datasets consist of images cropped to a bounding box carefully positioned around the respective teeth extracted from complete orthopantomogram (OPG) images. The datasets are independently used for dental stage classification. A pretrained DenseNet-201 model was fine-tuned on the datasets as a performance baseline. ViT models were also trained on the same datasets in order to establish an attention map baseline. Then we trained our proposed framework consisting of an autoencoder (AE) followed by a ViT model, with the end goal of analyzing the latent space representation of images, along with the differences in reconstructions and resulting ViT attention maps. All models were trained on a single NVIDIA A100 GPU with 80 GB of memory. The full training for a single fold, including the autoencoder and the subsequent classifier, took approximately 2 hours to complete. The trainings for both the DenseNet-201 and the ViT models on their own both took approximately 1 hour per fold. We then discussed our findings to understand the behavior of the trained models better, leveraging the increased transparency provided by image embeddings, the reconstructions, and the final attention maps. All these steps are expanded on in the remainder of this section. An overview of our proposed AE + ViT pipeline can be seen in Fig. \ref{fig:overview}.

\subsection*{Data}

\begin{figure}
    \centering
    \includegraphics[width=0.8\textwidth]{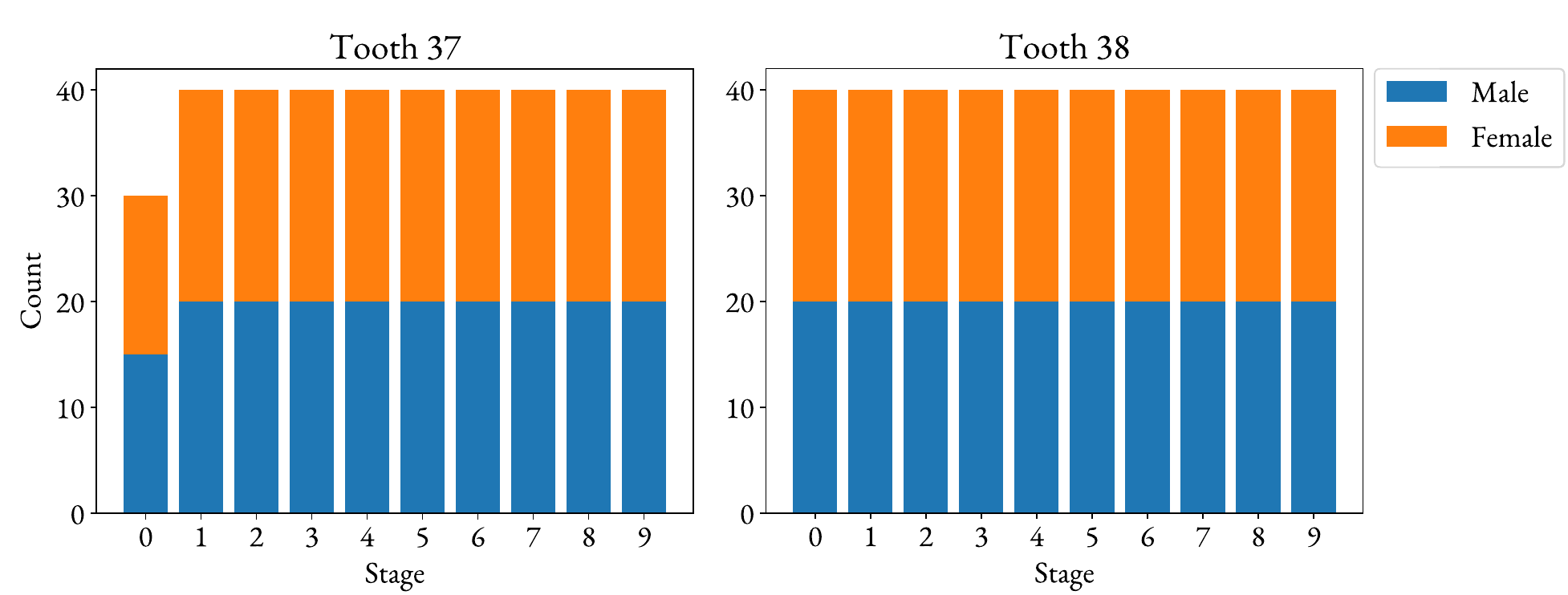}
    \caption{The number of samples per stage in the two datasets. The samples are balanced in stage and in sex, eliminating the
    need for methods which rectify class imbalance, and facilitating sex-invariant training.}
    \label{fig:stageDistrib}
\end{figure}

\begin{figure}
    \centering
    \includegraphics[width=\textwidth]{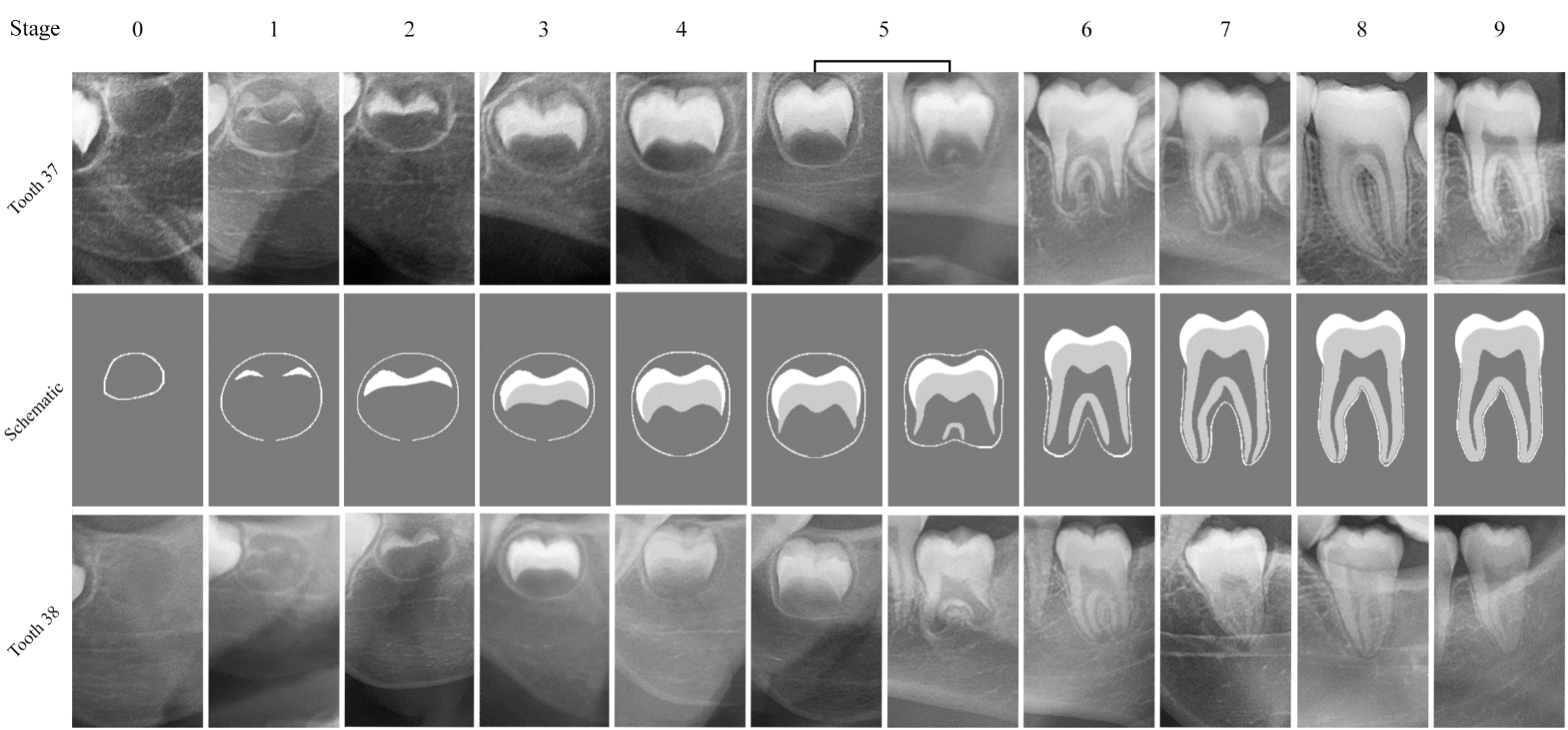}
    \caption{Schematic representations and illustrative case examples of the developmental stages, based on the modified Demirjian staging technique. Note the subtle differences in the bounding box sizes surrounding teeth 37 and 38. Tooth 38 appears relatively smaller than tooth 37 due to the use of larger bounding boxes, which were deliberately sized to fully encompass each tooth. This approach accounts for the greater anatomical and positional variability typically observed in tooth 38 compared to tooth 37.}
    \label{fig:stageSchematics}
\end{figure}

We utilise two separate datasets, one for tooth 37 and one for 38. The original OPGs were retrospectively selected from a set of 4000 OPG images collected at UZ Leuven, Belgium between 2000 and 2015, and subsequently anonymised by removing all patient information except for sex and age. Local ethics approval was obtained from the Ethical Commission Research UZ/KU Leuven (S62392), and consequently, informed consent was waived by Ethical Commission Research UZ/KU Leuven. The study was conducted in compliance with ICH-GCP guidelines, and the principles outlined in the World Medical Association Declaration of Helsinki on medical research. A primary selection pass was performed in order to exclude the records with (1) full absence of any permanent tooth type, (2) presence of orthodontic appliances, (3) bad image quality, (4) drastic overlap between any two teeth, and (5) pathologically abnormal teeth positions. The subjects were of Belgian origin within the age range of 0 to 24 years old, with a mean age of 14.56 $\pm$ 5.78. A secondary selection step resulted in 20 samples per stage, per tooth. Fig. \ref{fig:stageDistrib}. shows the number of samples per stage for the two teeth. The individual teeth were then cropped out of the original images with a standardised bounding box setup using Adobe Photoshop 2021\textregistered. The cropped images were labelled into ten developmental stages three junior investigator, who were trained and calibrated at our institution, and decided in consensus on the staging. A fourth senior investigator, with 19 years of experience in dental staging for age estimation, resolved disagreements. We chose to apply an augmented Demirjian scheme, described by De Tobel et al. (2017)\cite{de2017automated}. The scheme rates tooth development by assigning a stage number ranging from 0 to 9. The original Demirjian scheme ranges from stage A to stage H, with the advantage that it is based on anatomical criteria, whereas some other schemes are based on the predicted root length \cite{thevissen2013third}. In the augmented scheme, a stage 0 is added for the crypt stage, and a stage 8 for starting apical closure (which is intermediate between Demirjian G and H). This additional stage 8 is highly relevant around the age of 18 years old, and can therefore help increase the accuracy of discerning minors from adults \cite{thevissen2010human}. A visualization of teeth from each stage, and the diagrams for the stage criteria are shown in Fig. \ref{fig:stageSchematics}. The resulting tooth 37 dataset contained 390 tooth images (195 male, 195 female), and the tooth 38 dataset consisted of 400 images (200 male, 200 female). As for the preprocessing, all bounding box images were intensity-normalised individually, limiting the values to [0, 1] for all samples, and resized to 224x224 pixels to facilitate faster model training (Supplementary Fig. 1). In order to artificially expand the training data and improve model generalization, on-the-fly data augmentation was applied during all training steps. These augmentations included random brightness and contrast jitter with a probability of 0.3, and random affine transforms for all samples which consisted of random rotation in the range $[-5^\circ, 5^\circ]$, random translation by $[-12, 12]$ pixels in both the x-axis and the y-axis, and a random scaling by a factor of $[0.8, 1.2]$.  

The datasets were split into non-overlapping training, validation, and test splits for 4-fold cross-validation. The number of folds were chosen to ensure enough variation in the test set for each of the stages, with the additional reason of reducing the total training time. Per fold, 25\% of the dataset was assigned as the hold-out test set, 65\% as the training set, and 10\% as the validation set. Each split was stratified, preserving the balanced distribution of stage and sex of the full datasets.

\subsection*{DenseNet-201}

To establish a performance baseline for our study, a pretrained DenseNet-201 model, trained on the ImageNet dataset, was selected and fine-tuned. The DenseNet-201 architecture was previously used by Matthijs et al. (2024) to achieve 0.71 and 0.57 accuracy, respectively, for tooth 37 and 38, and was employed in this study due to its architectural efficiency and feature reuse. The foundational principle of the DenseNet family of models is their unique connectivity patterns, which seeks to maximize information flow between layers. Unlike traditional sequential architectures, each layer in a DenseNet model receives the feature maps from all preceding layers, and passes its own feature map to subsequent layers. Within a single block of the DenseNet-201 model, the output of the $l^{th}$ layer is computed as

\begin{equation}
    x_l = H_l\left([x_0,x_1,..., x_{l-1}] \right)
\end{equation}

where $\left([x_0,x_1,..., x_{l-1}] \right)$ represents the concentration of the feature maps produced in previous layers, and $H_l(\cdot)$ is a composite function of batch normalisation, ReLU activations and a $3\times3$ convolution. This dense connectivity offers several advantages. The enhanced feature reuse allows for the propagation of the low-level features learned in the earlier layers deeper into the network, allowing for more holistic decision making. The improved gradient flow achieved via the direct connections across layers mitigates the vanishing gradients problem often seen with deep neural networks. DenseNet models also commonly require significantly fewer parameters than other architectures to achieve similar performance, due to the dense connectivity encouraging feature reuse, which reduces model complexity and the risk of overfitting. This last advantage is particularly desirable in our application, as overfitting is a often a critical consideration when training deep models on datasets of limited size, such as ours, with the number of total samples is 390 and 400 respectively. 
Since the pretrained DenseNet-201 is designed to perform image classification with 1000 classes originally, a modification is required for our application. We append a final classification layer $H_{final} : \mathbb{R}^{1000} \rightarrow \mathbb{R}^{10}$ to the original model to adapt for 10 classes. 

In our experiments, we utilise the PyTorch implementation of the DenseNet-201 architecture, fine-tuned using the AdamW optimiser with a learning rate of \num{1e-4}. A weight decay factor of \num{1e-5} was employed to further reduce the risk of overfitting, alongside the use of a dropout ($p = 0.3$) layer before the final layer of the model. The models were trained to optimize for the cross-entropy loss. A learning rate scheduler was used, which, upon encountering a plateau in the validation loss, reduced the learning rate by a factor of 0.5, with a patience of 10 epochs. The models were trained with a batch size of 64, for a total of 300 epochs. 

\subsection*{ViT architecture}

In addition to the convolutional approach represented by the DenseNet-201, we investigated the performance of the ViT architecture. ViT models represent a paradigm shift from the convolutional models, adapting the Transformer model originally designed for natural language processing. The ViT works by slicing an input image into a contiguous, non-overlapping sequence of N patches,  where $N= (H\times W)/P^2$, with each patch having the size $P \times P$. These 2D patches are flattened into a 1-D vector representation and embedded to a lower-dimensional space via linear projection, creating patch embeddings. A class token \textit{CLS} is prepended to these embeddings. This token is a learnable parameter and acts as the main information and attention aggregator. A learnable position encoding is also added to these projections, allowing the model to keep track of the original position of each patch in the input image. This augmented linearised patch is then fed to the ViT encoder, which has the same structure as the original transformer encoder proposed by Vaswani et al. (2017)\cite{vaswani2017attention}, visualized in Supplementary Fig. 2. 
In the ViT encoder, the attention mechanism is a key component that enables the model to capture relationships between different parts of the input image. This mechanism allows each patch embedding to attend to all other patch embeddings, learning which parts of the image are most relevant to each other. The attention mechanism computes self-attention scores between pairs of patch embeddings by calculating how predictive each patch is of itself and other patches. These scores are then used to weight the importance of each patch when computing the output embeddings, by inspecting the attention of the \textit{CLS} token. Patches with higher attention scores contribute more to the final representation, allowing the model to focus on relevant regions and features within the image.

This attention system is stacked horizontally $M$ times to create a multi-headed attention module. Each attention head independently computes attention scores between pairs of patch embeddings, capturing different aspects of the image's spatial and semantic relationships. After computing attention scores, the outputs of the attention heads are concatenated to produce intermediate embeddings. By employing multiple attention heads, the model can attend to various features and patterns simultaneously, enabling it to capture both local and global dependencies more effectively. Additionally, this allows the model to learn richer and more nuanced representations of the input image. 
The multi-headed attention module, used in tandem with residual connections and a learnable mapping function, results in a single encoder layer. In a ViT encoder module, $L$ encoder blocks are vertically stacked, each layer learning the representations of the output of the previous layer. Using multiple encoder blocks in a visual encoder allows the model to learn rich and hierarchical representations of the input image, leading to improved performance and robustness.
After passing through the self-attention layers, the output corresponding to the \textit{CLS} token encapsulates the global context and features of the entire image. This aggregated representation can then be used for various downstream tasks, such as image classification. Here, we use a simple linear layer which takes the learned token vector and maps it onto the 10-dimensional stage predictions. In essence, the \textit{CLS} token acts as a global pooling mechanism, allowing the model to make predictions or decisions based on the holistic understanding of the input image, in addition to the localized information captured by individual patch embeddings.

In our experiments, we used $M=16$ attention heads, with $L=12$ encoder layers in the encoder block, with patch size $P=32$. We used the PyTorch framework to implement the ViT model. The training setup (learning rate, weight decay and dropout parameters) was kept exactly the same as the process used for the DenseNet-201 models, stated in the previous subsection, in order to establish a fair comparison between the two models.

\subsection*
{Autoencoder Architecture}

The classical AE setup is well-defined in the literature \cite{LOPEZPINAYA2020193}. This family of models embeds high-dimensional inputs to a low-dimensional latent space and, subsequently, decodes these embeddings for various purposes, such as reconstruction, inpainting, or denoising. They are also a valuable family of models to learn a latent space that displays desirable traits, a process that can be tuned by the selected loss function. 
We employ an AE architecture to reconstruct images removing image noise, reducing image-level intra-class variability and generating a latent space on which a data-centric contrastive investigation can be carried out.

The used convolutional AE is a two-stage architecture, consisting of an encoder and a decoder. The encoder network $f_{enc}$ maps an input image $I$ to a latent vector $z \in \mathbb{R}^{32}$. This transformation is achieved through a composition of five sequential convolution blocks,

\begin{equation}
    z = f_{enc}(I)  = (f_{linear} \circ \text{Dropout}(p=0.3) \circ f_5 \circ \cdots \circ f_1)(I)
\end{equation}

where $f_i, i\in{[1,2,\cdots,5]}$ signifies a $3 \times 3$ convolution with a stride of 2, followed by a ReLU activation $\sigma_{R}$. Hence a single block can be described as $X^k = \sigma_{R}(W^k \star X^{k-1}+b^k)$, where $X^{k-1}$ is the input feature map, and $W^k$ and $b^k$ are learnable filter weights and biases. This encoder structure downsamples the spatial resolution while doubling the feature channel depth at each stage. The final convolution layer is followed by a dropout operation to mitigate overfitting. The final feature vector is flattened and projected to the latent space via a dense linear layer $f_{linear}$. 

The decoder network $g_{dec}$ generates a reconstructed image $\hat{I} = g_{dec}(z)$ from the latent vector $z$. To preclude the emergence of "checkerboard artifacts" commonly seen when using transposed convolutions, our decoder network decouples the upsampling and convolution operations. The decoder architecture symmetrically reverses the encoder operations, using bicubic sampling with a factor of 2 followed by a $3 \times 3$ convolutional layer for feature refinement at each of its five blocks. The final layer employs the sigmoid activation function $\sigma_{S}$ to ensure the output $\hat{I}$ is a valid probabilistic map with pixel intensities in the range $[0,1]$.

In order to shape the learned latent space, and the visual quality of the image reconstruction, all instances of the AE architecture were trained via end-to-end backpropagation using a multi-component objective function. This function was engineered to concurrently satisfy two distinct objectives; high-fidelity probablistic image preprocessing and ordinal metric learning in the latent space. The total loss $L_{total}$ is a linearly weighted combination of two components,

\begin{equation}
    L_{total} = \gamma \cdot L_[triplet] + (1 - \gamma) \cdot L_{recons}
\end{equation}

where $L_{triplet}$ is a modified triplet margin loss\cite{chen2017beyond,cheng2016person} with a variable margin, and $L_{recons}$ is a composite loss function consisting of the binary cross entropy (BCE) loss and the Learned Perceptual Image Patch Similarity (LPIPS) loss\cite{zhang2018unreasonable}. $L_{triplet}$ can be formulated as,

\begin{equation}
    \begin{split}
        L_{triplet} &= \max\left(0, D(z_a^{norm}, z_p^{norm}) - D(z_a^{norm}, z_n^{norm}) + \alpha_{ordinal}\right)\\
        D(z_1, z_2) &= \sqrt{\sum_i{(z_{1,i} - z_{2,i})}^2}\\
        \alpha_{ordinal} &= \left\lVert {y_a - y_n}/9 \right\rVert
    \end{split}
\end{equation}

where $D(\cdot)$ is the Euclidean distance, $z^{norm}$ is the L2 norm of the embedding vector $z$, $y_{a}, y_{n} \in [0,1,\cdots,9]$ stand for the stage labels of the anchors and the negative samples and $\alpha_{ordinal}$ is the variable margin value. The $\alpha_{ordinal}$ margin is the main modification to the training strategy. Given two negative samples of consecutive stages, the margin, or the minimum distance between the embeddings that satisfies the triplet criterion, equals 0.1, whereas when the negative sample is further apart from the anchor, the margin grows up to a value of 1. This variations in the margin depending on the negative stage label allows the AE model to consider the chronological order of developmental stages in the latent space by pushing chronologically closer samples closer together that those with larger developmental distance. This effect results in reconstructions that are more similar for similar stages while preserving inter-class variability. To further ensure stable learning, semi-hard triplet mining is used\cite{chung2023joint}.

The reconstruction loss $L_{recons} = L_{BCE} + L_{LPIPS}$ is the sum of the two functions $L_{BCE}$ and $L_{LPIPS}$, which are defined as in equations \ref{eq:bce} and \ref{eq:lpips}.

\begin{equation}\label{eq:bce}
    L_{BCE}(I,\hat{I}) = -\frac{1}{H\cdot W}\sum_{i=1}^{H}\sum_{j=1}^{W}\left[ I_{ij}\log(\hat{I}_{ij}) + (1-I_{ij})\log(1-\hat{I}_{ij}) \right]
\end{equation}

In Eq. \ref{eq:bce} $H,W$ are the height and width of the input image $I$, $\hat{I}$ is the reconstructed image. By treating the images as probabilistic maps where all pixel values are between 0 and 1, the BCE loss function promotes sharper reconstructions than other conventional losses such as the mean squared error. As such, the BCE loss plays a part in creating human-readable images.

\begin{equation}\label{eq:lpips}
    L_{LPIPS}(I,\hat{I}) = \sum_{l\in L} \frac{1}{H_l W_l}\sum_{h=1}^{H_l}\sum_{w=1}^{W_l} \norm{w_l \odot (F^l_{hw}(I) - F^l_{hw}(\hat{I}))}^2_2
\end{equation}

The LPIPS loss is an image similarity metric that computes the distance between deep feature representations of the ground-truth and generated images, extracted from a pretrained VGG-16 network. This loss metric tends to align closely with human perception, moreso than metrics that focus on pixel similarity\cite{zhang2018unreasonable}. In Eq. \ref{eq:lpips}, $F^l$ stands for the channel-wise normalised activation maps from the layer $l \in L$ of the VGG-16 model, with $(h, w)$ are the spatial positions in layer $l$, and $w_l$ is the calibrated channel-wise weight, fine-tuned by Zhang et al. (2018) \cite{zhang2018unreasonable}. By minimising the distance between the deep feature representations during training, the AE is motivated to produce reconstruction images that not only appear similar to the input images, but also trigger similar feature filters to the inputs. This loss function, therefore, motivates reconstructions that are semantically similar to input images.

In our experiments, we used $\gamma = 0.7$, weighing the triplet loss more heavily in order to ensure the semantic structure of the latent space, while also constraining the reconstructions to be visually similar to input according to the human eye. The AdamW optimiser was utilised for all AE training, with the learning rate of \num{5e-4} and a weight decay factor of \num{1e-5}. The training was carried out for 300 epochs with a batch size of 128. The validation loss was monitored for convergence for early stopping, which did not take place during the 300 epoch period in any of the training folds. The loss curves from all folds are displayed in Supplementary Fig. 7 and 8.

\subsection*{Evaluation Metrics}
To quantitatively assess the model performances, a suite of standard evaluation metrics was employed. Given the class-wise balanced classification task we perform, the accuracy score (Eq. \ref{eq:acc}) was a suitable metric.

\begin{equation}\label{eq:acc}
    \text{Accuracy} = \frac{\text{Number of Correct Predictions}}{\text{Total Number of Predictions}}
\end{equation}

The accuracy score shows the rank-1 recognition rate, however, in the case of ordinal stages, it may be misleadingly pessimistic. Since a prediction that is one stage off is objectively a better one than one that is off multiple stages, other metric should also be reported to holistically demonstrate the model performance. To this end, we also report the linearly weighted Cohen's $\kappa$ score (Eq. \ref{eq:kappa}). This metric extends the standard kappa statistic by applying weights to quantify the degree of disagreement between predictions and true labels, thereby penalizing large errors more severely than minor ones. The weighing scheme is directly proportional to the absolute difference between the true class and the predicted class. This ensures that large errors (e.g. predicting stage 0 when the true stage is 5) are penalized more heavily than minor errors (e.g. predicting stage 4 for the true stage 5).

\begin{equation}\label{eq:kappa}
    \kappa_\omega = \frac{p_o - p_e}{1- p_e}
\end{equation}

In Eq. \ref{eq:kappa}, the observed agreement $p_o$ and the chance agreement $p_e$ are computed as weighted averages based on this linear disagreement scale. Finally, in order to evaluate the retrospective regression performance of the models, the conventional metric of mean absolute error (MAE) is utilised.

\begin{equation}
    MAE = \frac{1}{n}\sum_{i=1}^{n}\left\lVert y_i - \hat{y}_i \right\rVert
\end{equation}

As the MAE does not square the errors, it is a depiction that is less sensitive to large outliers, and it has the advantage of being expressed in the same units as the target variable, facilitating direct interpretation of the magnitude of errors.

\section*{Results and Discussion}

\begin{table}[!htb]
\centering
\caption{The cross-validation metrics of the staging performance of all evaluated methods. The best metrics for each tooth are reported as bold text. All metrics are presented as mean (std). }
\label{tab:vit_res}

\begin{tabular}{c|ccc|ccc}
\multirow{2}{*}{\textbf{Method}} & \multicolumn{3}{c|}{\textbf{Tooth 37}}                                                      & \multicolumn{3}{c}{\textbf{Tooth 38}}                                                       \\ \cline{2-7} 
                                 & \multicolumn{1}{c|}{\textbf{Accuracy}} & \multicolumn{1}{c|}{\textbf{MAE}} & \textbf{Kappa} & \multicolumn{1}{c|}{\textbf{Accuracy}} & \multicolumn{1}{c|}{\textbf{MAE}} & \textbf{Kappa} \\ \hline
\textbf{ViT Only}                & 0.712 (0.025)                          & 0.375 (0.026)                     & 0.680 (0.028)  & 0.462 (0.020)                          & 0.867 (0.054)                     & 0.402 (0.022)  \\ \hline
\textbf{AE + ViT}                & \textbf{0.815 (0.022)}                          & 0.252 (0.025)                     & \textbf{0.794 (0.033)}  & \textbf{0.543 (0.052)}                          & 0.711 (0.051)                     & \textbf{0.492 (0.058)}  \\ \hline
\textbf{DenseNet Only}           & 0.810 (0.030)                          & \textbf{0.216 (0.065)}                     & 0.788 (0.034)  & 0.535 (0.066)                          & \textbf{0.679 (0.201)}                     & 0.483 (0.073)  \\ \hline
\textbf{AE + DenseNet}           & 0.748 (0.023)                          & 0.314 (0.021)                     & 0.720 (0.026)  & 0.485 (0.049)                          & 0.792 (0.085)                     & 0.427 (0.054)  \\ \hline
\end{tabular}
\end{table}

We first present our findings on the direct staging of dental images for teeth 37 and 38 using the ViT and DenseNet-201 models as a performance baseline and establish a comparison of the two classifier models. The statistical significance of the results are presented in Supplementary Table 2.

\begin{figure}
    \centering
    \includegraphics[width=\textwidth]{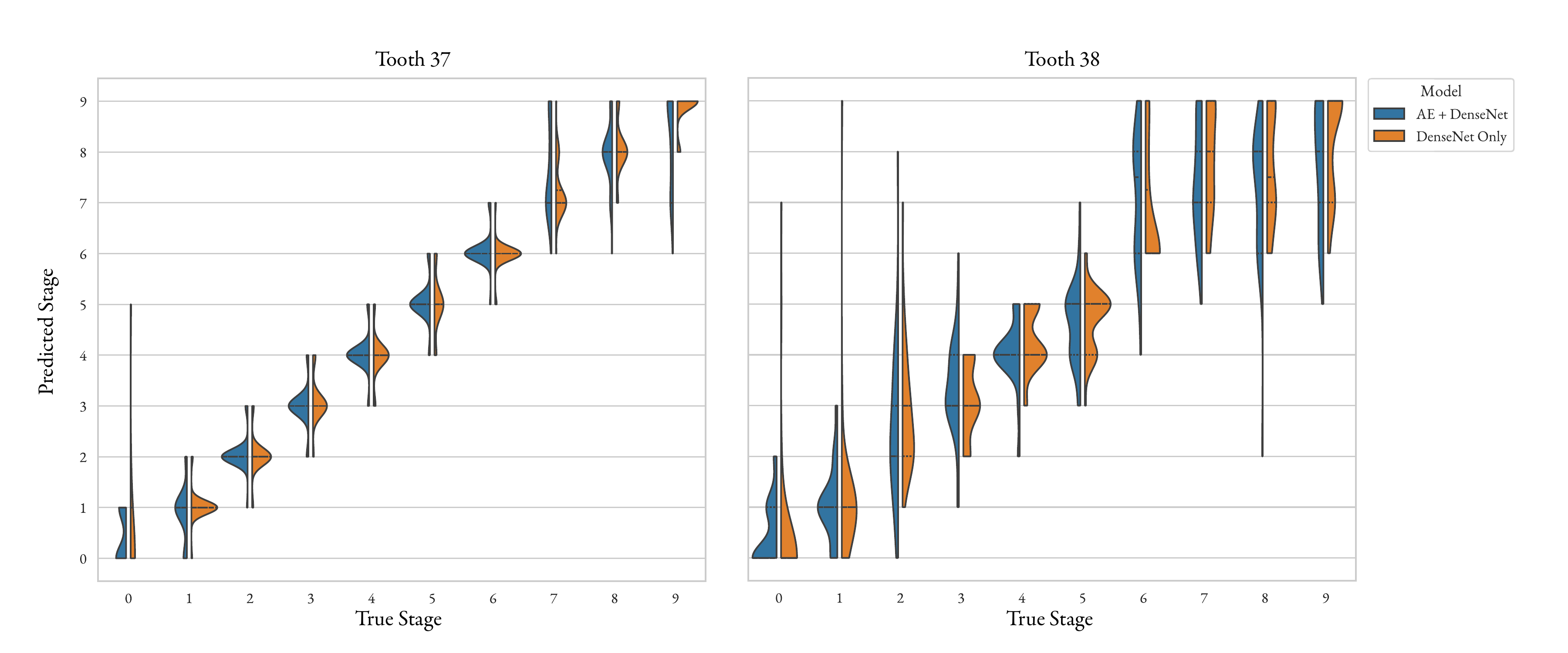}
    \caption{The violin plots for illustrating the distribution of predicted dental stages for two different models, the DenseNet-201 only (orange) and AE + DenseNet (blue). The analysis is presented separately for tooth 37 and 38. For tooth 37, both models show a strong correleation between the annotated stage and the predictions, with the AE + DenseNet model showing more error in the latest stages of 7, 8, and 9. Conversely, for tooth 38, both models exhibit a larger error margin. The DenseNet-201 predictions for tooth 38 are slightly more clustered around the ground truth labels than those of the AE + DenseNet model. }
    \label{fig:violinsDense}
\end{figure}

The DenseNet-201 model was trained on the original images after preprocessing via 4-fold cross validation. The performance metrics, shown in Table \ref{tab:vit_res}, show that the DenseNet-201 architecture reached and an accuracy of 0.810 and an MAE of 0.216 for tooth 37, and 0.535 accuracy with an MAE of 0.679 for tooth 38. This indicates the DenseNet-201 models, across all folds, were much more successful in agreeing with the original annotations for tooth 37 than for tooth 38. Furthermore, the violin plots of the DenseNet-201 predictions versus the ground truth labels can be seen in Fig. \ref{fig:violinsDense}. These predictions are collected from each fold in order to represent the performance of the DenseNet-201 across the entire dataset. The predictions for tooth 37 are on average one stage off, with outliers from stages 0 and 7. For tooth 38, the predictions are more dispersed, showing the inability of the DenseNet-201 model to find distinguishing features per stage annotation. 

\begin{figure}[!ht]
    \centering
    \includegraphics[width=\textwidth]{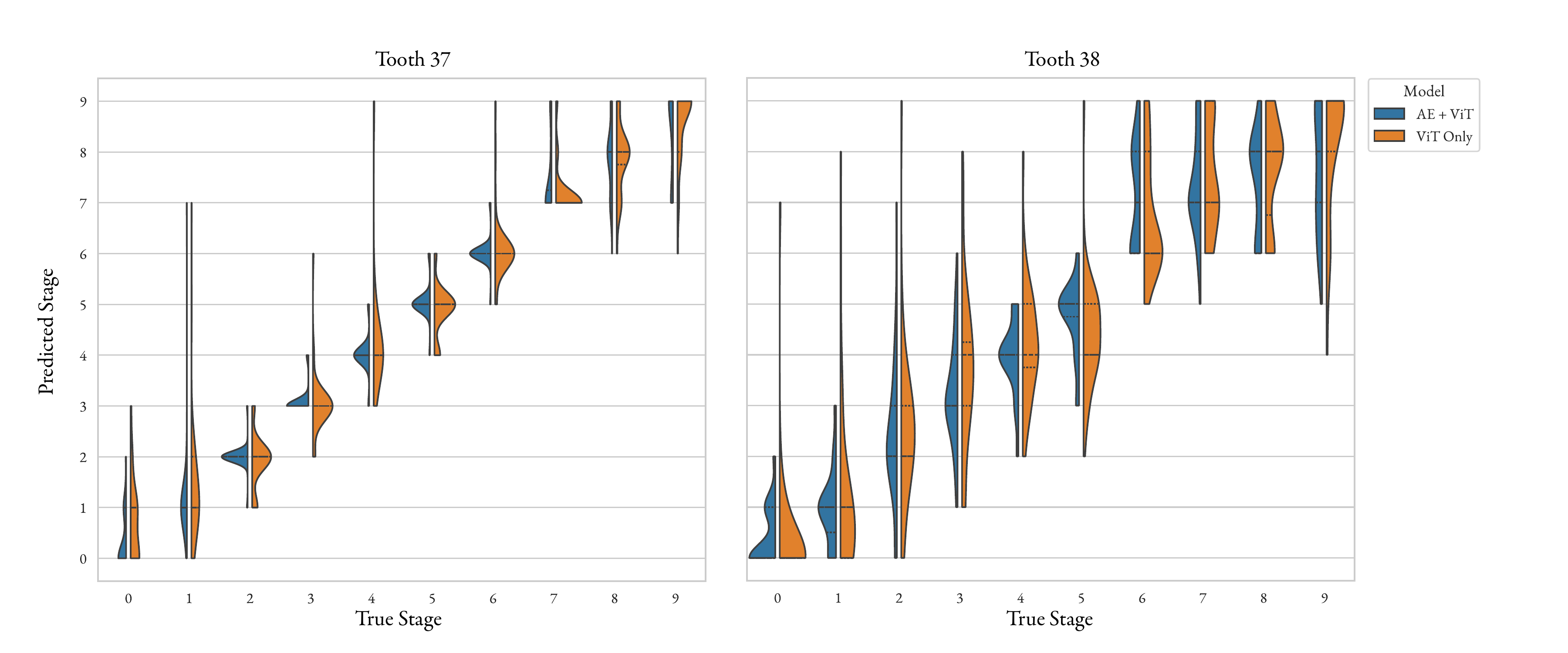}
    \caption{The distributions of ViT stage predictions versus ground truth labels per stage, from the AE + ViT framework and using the ViT model only without image reconstruction. The predictions are acquired via cross-validation. It can be seen that the predictions for tooth 37 are much more concentrated on the true labels, compared to those of tooth 38, regardless of the use of an AE model. For both teeth, the predictions became more concentrated on the ground truth when the AE model is utilised, indicating that the reconstruction step makes classification easier for the ViT model.}
    \label{fig:violinsVit}
\end{figure}

Additionally to the DenseNet-201, the ViT model is also trained directly on the original preprocessed images, in 4-fold cross-validation fashion. The results can be seen in Table \ref{tab:vit_res}. These results coincide with the large difference in automated staging performances on tooth 37 and tooth 38 observed with the DenseNet-201, with the latter tooth displaying less favourable metrics. We can further see the difference in predictive performance in Fig. \ref{fig:violinsVit}, where all the ViT predictions from each fold in the cross-validation are visualized. For tooth 37, most false predictions through stages 2 to 8 lie within a one-stage error margin, with predictions for stages 0 and 9 displaying errors on the extreme side (e.g. predicted 5 with label 0). However, the means show the predictions are still concentrated on the correct target. Conversely, the smallest error distribution for tooth 38 is seen in stages 7 and 8, spanning three stages. It is clear from these plots that the ViT model for tooth 37, approaching the ground truth label with few outliers seen in stages 0, 1, and 4, generally approaches the desired stage prediction, while tooth 38 does not display the same behaviour.
This is an unfavorable result from a forensic odontology point of view, as the development of tooth 38 is regarded as a good proxy for chronological age in young people, since it develops through the ages of 7 to 21 \cite{liversidge2008timing}. Moreover, it is recommended to assess third molar development alongside the development of hand/wrist bones and clavicles for age estimation in adolescents and young adults \cite{schmeling2016forensic}.

In order to further analyze this performance disparity in the classification performance of the two molars, encountered with both the ViT and the DenseNet-201 models across all folds, we apply our proposed framework. As seen in Fig. \ref{fig:overview}, this framework consists of an AE model, trained to learn a meaningful latent space while reconstructing images as categorical prototypes by representing the distinguishing visual properties of each stage while also incorporating sample-based differences. Following the AE preprocessing, the images are passed to the classification model. In our proposed method, we utilise the ViT as the primary classification model due to its attention mechanism, while also reporting on the version of the framework using DenseNet-201 as the classifier model, thus demonstrating the effects of the classification model on the final performance. However, we exclude the framework with the DenseNet-201 while discussing the explanations, as this version is unable to offer attention maps. 

%

    Based on the evaluation metrics presented in Table \ref{tab:vit_res}, the inclusion of the AE model as a preprocessing step is associated with improved performance for the ViT architecture across both tooth datasets. For tooth 37, the AE + ViT pipeline yielded an accuracy of 0.815 and an MAE of 0.252, compared to an accuracy of 0.712 and an MAE of 0.375 for the ViT model only. This trend was also observed for the tooth 38 dataset, where the AE + ViT model resulted in an accuracy of 0.543 and an MAE of 0.711, relative to the ViT model's accuracy of 0.462 and MAE of 0.867. These results suggest that the AE's function of reconstructing images may reduce input "noise" and intra-class feature variability, creating a representation that is more readily classifiable by the ViT architecture, displaying the stage-specific characteristics in the reconstructions. A similar enhancement can be observed in the agreement of the predictions with the ground truth annotations with the linearly weighted Cohen's $\kappa$ score rising from 0.680 to 0.794.
    On the other hand, the use of AE-based preprocessing corresponds to a decrease in performance for the DenseNet-201 architecture on both datasets. For tooth 37, the DenseNet only approach produced an accuracy of 0.810 and an MAE of 0.216. With the AE prefix, the accuracy was reduced to 0.748, and the MAE increased to 0.314. This pattern was consistent for tooth 38 as well, with the accuracy for the AE + DenseNet model being lower than that of the DenseNet-201 model, with an accuracy of 0.485 and MAE of 0.535. One noticeable detail was the high variation in MAE of the DenseNet-201 model for tooth 38 being quite high compared to the other metrics, indicating the DenseNet only approach may be less stable across folds.

\begin{figure}[!htb]
    \centering
    \includegraphics[width=\textwidth]{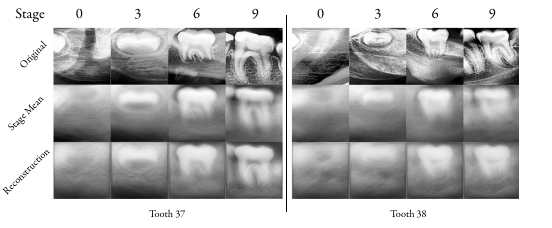}
    \caption{Examples of original samples and AE reconstructions, contrasted against mean stage images for tooth 37 and tooth 38. Tooth 37 reconstructions show a denoising effect while reframing images as stage prototypes, and offer readable reconstructions which are sharper than the stage mean image, and depict the dental structures more clearly. Tooth 38 reconstructions, while having similar features, depict much blurrier prototypes, which both causes, and indicates to the cause of, the low classification accuracy, namely large intra-class variability. Even with their blurrier nature, the reconstructed images are still sharper than stage mean images. This shows, for both teeth, that the AE model does more than simply representing an average tooth per stage, and instead encodes the images with the anatomical structures characteristic to the stage, making classification easier for the downstream models. }
    \label{fig:ims_recons}
\end{figure}

    The divergent outcomes for the ViT and DenseNet architectures may be attributed to their distinct operational principles. The AE pipeline is engineered to reduce image noise and generate smoothed, prototypical reconstructions for each developmental stage (Fig. \ref{fig:ims_recons}). This more abstract representation aligns well with the ViT architecture, which processes images by learning global spatial relations between patches, rather than focusing on high-frequency details. By simplifying the input, the AE appears to enable the ViT's self-attention mechanism to focus more effectively on diagnostically relevant anatomical structures. In contrast, the DenseNet-201 model, as a Convolutional Neural Network, leverages a hierarchy of features, including the low-level textural information propagated through its dense layers. The smoothing effect of the AE may degrade these fine-grained features, thereby hindering the model's discriminative capability. Furthermore, since the DenseNet model was pretrained on the ImageNet dataset, its feature extractors are optimized for complex natural images. The simplified, prototypical outputs from the AE may introduce a domain mismatch, leading to less effective feature extraction and a subsequent reduction in classification accuracy.

Since the AE+ViT pipeline can be probed for information by design, a contrastive inspection can be carried out between the two teeth starting with an inspection of the AE preprocessing. A visual inspection of the AE effect was performed by comparing original images, their corresponding reconstructions, and the mean image for each stage, as depicted in Figure \ref{fig:ims_recons}. This analysis reveals that the AE architecture imparts a smoothing effect on the inputs. The resulting reconstructions are distinct from simple stage-mean images; while they are guided towards a stage prototype, they retain sample-specific variations. This effect stems from the loss functions employed during AE training. The triplet condition dictates that images of the same stage are encoded close together, meaning their reconstructions start from the same region in the latent space. This resulted in the reconstructed images being more similar if they share a stage label. The stochastic differences in images, which were crucial to avoid generating the same output for all images from the same stage, were incorporated using the BCE and LPIPS losses, enforcing the reconstructions to be similar to the inputs. This effect is shown in Supplementary Fig. 3.

For tooth 37, AE preprocessing yields clear prototypes with reduced noise, where dental structures are depicted more clearly than in the mean stage images. In contrast, the reconstructions for tooth 38 exhibit considerable blurring, particularly in the root and crown regions, and deviate more from the original images. These less coherent visual features for tooth 38 suggest difficulty in forming distinct stage prototypes, likely reflecting the high intra-class variation in the dataset, and may be a contributing factor to the lower classification accuracy reported for this tooth.

\subsection*{Comparison of Attention}

Through leveraging the attention mechanism of the ViT classifier, attention maps can be created to localize the most impactful areas of the input images on the class decision. We visualize the learned attention of the model using the attention rollout method\cite{abnar2020quantifying}. Comparing the attention maps of the ViT-only approach with the AE+ViT classification for teeth 37 and 38 can help understand how regions of focus change for each approach.  

\begin{figure}[!htb]
    \centering
    \includegraphics[width=\textwidth]{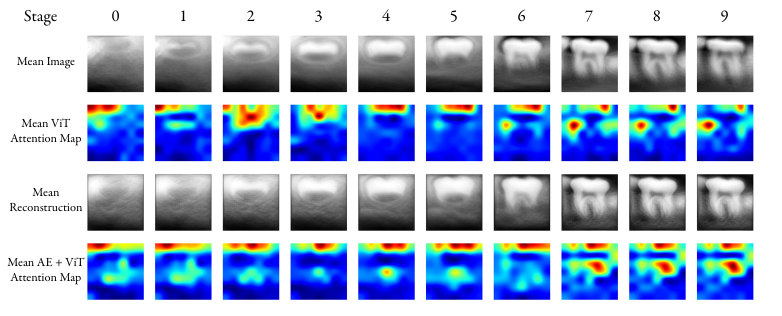}
    \caption{
    Comparison of the mean stage images and attention maps for tooth 37. Top row: Mean stage images, a general representation of stage shape. Second row: Mean ViT attention maps, showing how the ViT focus changes across stages by moving from the tooth cusp to the cervical section as stages progress. Third row: Mean reconstructions per stage, depicting how, on average, the reconstructions represent the stage shape. Bottom row: The attention maps from the AE+ViT framework. It can be seen that the attention maps for the ViT model heavily focus on the tooth cusps, and mostly disregard the root region, while the attention maps from the AE + ViT framework incorporate the root information much more, indicating that this useful anatomical feature was pronounced by the AE preprocessing.}
    \label{fig:vit_aevit_37}
\end{figure}

Figure \ref{fig:vit_aevit_37} provides insight into the decision process of the ViT model with and without the AE prefix. For the baseline ViT-only model, the mean attention maps display heavy focus on the tooth cusps, moving towards the cervical region as development progresses. In the earlier stages, attention is mainly on the the peripheral elements such as the neighboring tooth and crypt formation. For the later stages (7-9), in which root development is close to completion, the model critically fails to shift the focus to the root region, an area that is essential in discriminating between these advanced developmental stages \cite{de2017automated}. In contrast, the AE + ViT model displays a more distributed attention pattern in the earlier stages, focusing on the aforementioned elements that the ViT model also attends to, but also showing heavy focus on the cervical and root regions. It is important to notice that the attention maps only localize the sections that were influential in the decision, but they do not indicate towards which direction the regions affected the final output. In this light, the attention on the root regions, even in stages that do not display root formation at all, is an indication that the model considered \textit{the lack of root formation}, instead of the existence of it. In the later stages, the attention shifts heavily toward the cervical and apical sections, contrasting the attentions of the ViT-only model. These observations indicate that the AE preprocessing pronounced useful anatomical features more clearly than the original images. The ability of the AE + ViT model to leverage visual information from the roots indicates that its decision-making aligns more closely with the diagnostic criteria of a human observer (Fig. \ref{fig:stageSchematics}) and may account for the improved accuracy over tooth 38.

\begin{figure}[!htb]
    \centering
    \includegraphics[width=\textwidth]{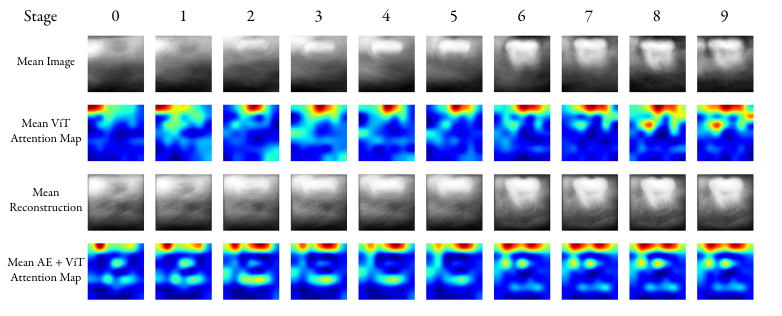}
    \caption{Comparison of mean stage images and attention maps for tooth 38. Top row: Mean stage images reveal a blurrier average per stage, indicating the tooth shapes for 38 contain more intra-class variation. Second row: Mean ViT attention maps seem similar to those of tooth 37, remaining plausible; however they do not incorporate the information below the mid-region of the tooth, and thus do not explain the lower accuracy. Third row: Mean reconstructions are less visually similar to mean stage images, indicating the mean images are not the optimal representation of stage morphology. Bottom row: The attention maps from the AE + ViT pipeline ,for all stages, focus on the lower region of the tooth more than ViT, indicating that the root formation informed the classification process.}
    \label{fig:vit_aevit_38}
\end{figure}

Upon inspecting the attention maps of the two methods applied to tooth 38 in Fig. \ref{fig:vit_aevit_38}, we can observe a similar focusing effect on the attention maps, induced by the AE preprocessing. The mean stage images for tooth 38 appear blurrier and less distinct compared to those of tooth 37. This implies a higher degree of intra-class variation in this lower-performing dataset. Especially seen in later stages of 6 to 9, the cervical to apical region of the roots in the mean stage images lack a distinct shape, hinting at the varied morphology of this area. The attention maps of the ViT only approach can be seen to mirror this lack of distinction, where the average attention values from stages up to 6 are of a spatially diffuse nature. This is a crucial observation for understanding the lower classification accuracy for this tooth. As discussed before, the ViT architecture relies on the global spatial relations between patches, which assumes the patches contain visually similar information (e.g. the roots being in similar positions for all images). The varied morphology of tooth 38, especially prevalent around the roots as shown by the mean stage images and Supplementary Fig. 6, fundamentally disagrees with this assumption and results in an unfavorable classification accuracy of 0.462. This phenomenon is echoed in the attention maps for the ViT-only approach with the mean attention maps showing focus only around the visually similar patches across all images of the same stage, and drastically differing from image to image. Proceeding to the AE + ViT framework, we can immediately observe the increased sharpness of the mean reconstructions, indicating a reduction in the visual variation within stages. The increased resolution observed especially around the roots in stages 6 to 9 indicates the AE model reduced the intra-class difference, allowing for easier classification. The mean attention maps of the AE + ViT pipeline reflect this by displaying much more focused attention patterns. From the earliest stages, attention 'hot-spots'. appear in the mid-section of the images, incorporating information from anatomically relevant regions in the decision process. Similarly to Fig. \ref{fig:vit_aevit_37}, we observe increased attention on the cervical and root regions of the teeth in stages 6-9. However, this increase is to a lower degree compared to the one seen for tooth 37. It can be thus concluded that even with the AE prefix, the pipeline was unable to learn the distinguishing features for the root area of the region, though still encoding useful information in those features that can account for the increased classification performance with the AE + ViT network.

\begin{figure}[!hp]
    \centering
    \includegraphics[width=\textwidth]{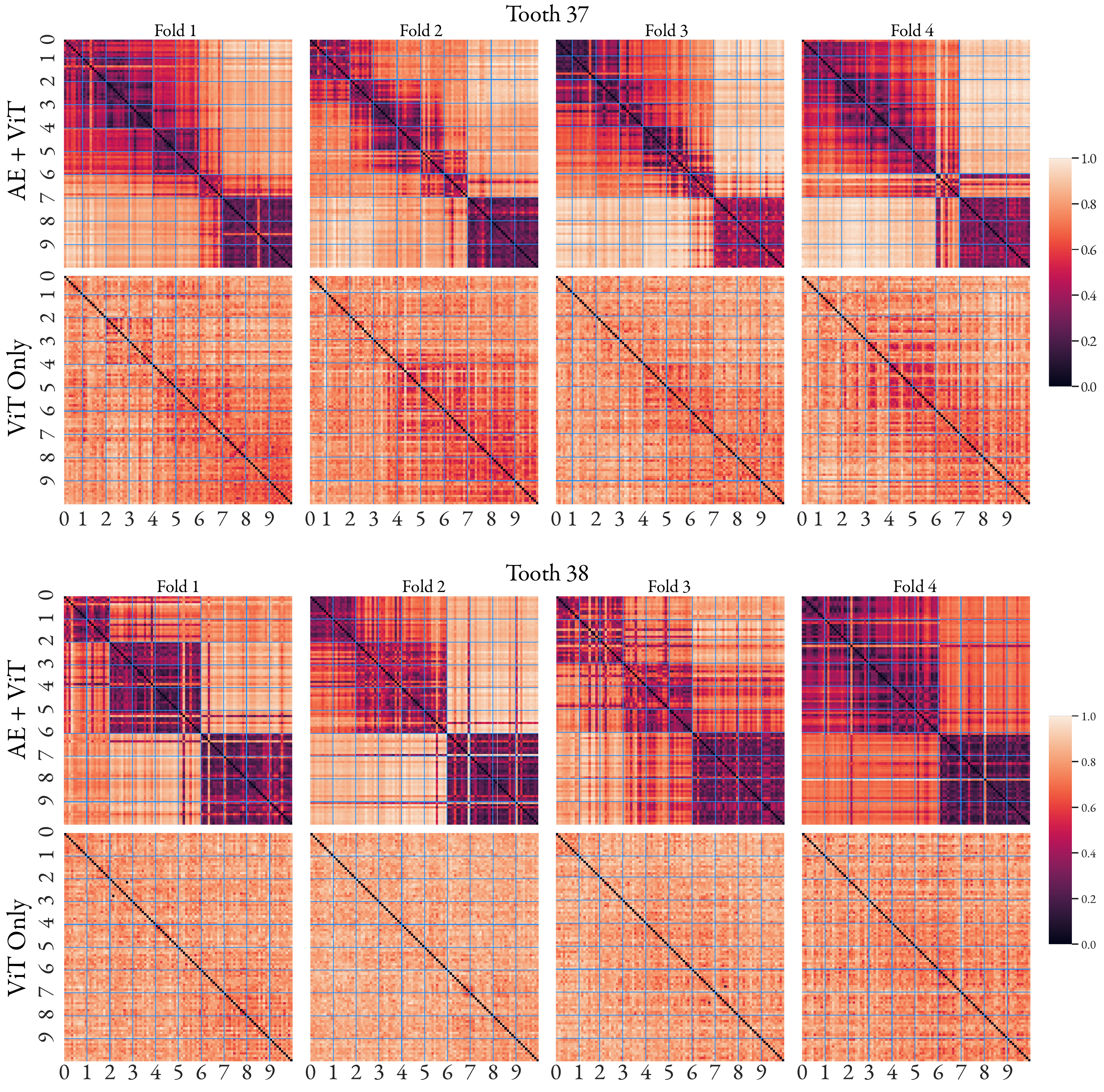}
    \caption{Heatmaps of LPIPS loss between the attention maps of each sample in the test set, per fold and model. The samples are ordered by stage label, and axis labels denote changes of label. Each fold is visualized independently. The LPIPS loss signifies perceptual similarity of attention maps, the lower the value the higher the similarity. When only the ViT model is used, the attention patterns are mostly different, with attentions of the same class uncorrelated. The attention maps of the AE + ViT pipeline show greater similarity, indicating the model was able to find specific patterns in similar locations. This is the intended effect of the AE prefix and it makes classification easier. }
    \label{fig:attSimHeatmaps}
\end{figure}

In order to further analyse how the difference between the performances of the ViT and AE + ViT models is reflected in the attention maps, a similarity inspection for all attention maps is beneficial. Fig. \ref{fig:attSimHeatmaps} shows the LPIPS loss between all attention maps from all folds, obtained using the ViT and the AE + ViT approaches. It is immediately noticeable that the similarities between the attention maps when using the ViT only is very low, meaning the ViT model does not localize the attention in any one region across all images for the same stage label. This is more apparent for tooth 38 than for tooth 37. For tooth 37, large blocks of smaller LPIPS loss can be seen in a distributed fashion (e.g. stages 4–9 in fold 2), but a similar structure is altogether lacking for tooth 38, where each attention map appears markedly different from the others. This indicates that the model is unable to identify similar structures in comparable locations to guide its decision-making, and must instead attend to each sample individually. As a result, the ViT model cannot effectively reuse the features it has already learned, but must adapt to each image separately.This, in turn, necessitates the use of substantially larger models with many more parameters in order to fit these datasets. We further support this argument with the classification performance achieved by the DenseNet-only approach. Since the DenseNet-201 architecture is a large model with highly efficient feature reuse, the DenseNet-only approach was able to yield a better accuracy than the ViT-only approach. Hence, the AE prefix shows an additional advantage of reducing the classification model size requirement. With AE preprocessing in the AE + ViT pipeline, for both tooth 37 and 38, Fig. \ref{fig:attSimHeatmaps} demonstrates a block-diagonal structure in the similarity heatmaps of the attention maps. For tooth 37, the earlier stages of 0 to 5 show similar attention maps, with sub-clusters of even higher similarity existing within this range. The stages 6 to 9 also show increased similarity for this tooth, with stages 7, 8 and 9 being highly similar. For tooth 38, the block-diagonal structure of the heatmaps suggests high similarity between stages 0 to 2, 3 to 6, and 7 to 9; however, this is variable across folds. Nevertheless, the increased similarity between the attention values between subsequent stages is an indication that the AE + ViT framework was able to reduce intra-class visual variation while to a degree preserving, or increasing inter-class variation. While this effect resulted in a performance increase and better attention maps, it is worth noting that in the ideal case, the dissimilarities between attention maps of samples of different stages would be greater. The prototyping function of the AE maps the input images closer to an ideal stage image. However, there is also the risk of having these reconstructions \textit{too similar}, especially for morphologically closer stages such as 7, 8 and 9. Due to a shared shape basis of these stages, it is understandable that the attention patterns tend to converge to a common basis. Nevertheless, caution is advisable in the analysis of the resulting attention patterns due to the risk of multi-stage converge. For this reason, our AE model training employed image-based and deep feature-based losses in order to avoid generating similar images for different stages for this reason.

An interesting division in attention map similarity exists for both teeth, between stages 0 to 5 and 6 to 9. This division overlaps with human understanding in the sense that root development is advancing from stage 6 onwards, while stages 0 to 5 only concern crown formation(Fig. \ref{fig:stageSchematics}). The division between attention maps at this threshold implies that the root formation---a feature that was disregarded without AE preprocessing---was highly influential in the decision process of the AE + ViT framework for automated staging, and that the inclusion of our specific AE architecture has helped the models approximate expert decision process better. 

These observations highlight the benefit of using an AE in the classification process along with the ViT classifier, namely the reduction of intra-class variation and a focus of attention more on the relevant regions. However, while the attention maps generated by ViT provide information about the localisation of attention, these maps differ only slightly across stages in this specialized forensic application, and are plausible, even though the classification performance is quite low. Based on this deduction, additional investigation into the latent space, made possible by the AE prefix, becomes desirable, which is the focus in the next section. For an example of how the attention evolves throughout the layers of the ViT, we refer the reader to Supplementary Fig. 4.

\subsection*{Latent Space Analysis}

To investigate the performance issues for tooth 38 beyond the visual level, we leverage the latent space of the AE, which we consider to be the main advantage of using an autoencoder in the classification pipeline. Since the latent space, induced by the variable margin triplet loss, has a metric property and the distances between the embeddings are meaningful, it is possible to inspect the encodings to draw conclusions regarding model behaviour. We employ principal component analysis (PCA) based linear dimensionality reduction to represent the 30-dimensional code in a 3-dimensional PC-plot for ease of visualization. 

\begin{figure}[!htb]
    \centering
    \includegraphics[width=\textwidth]{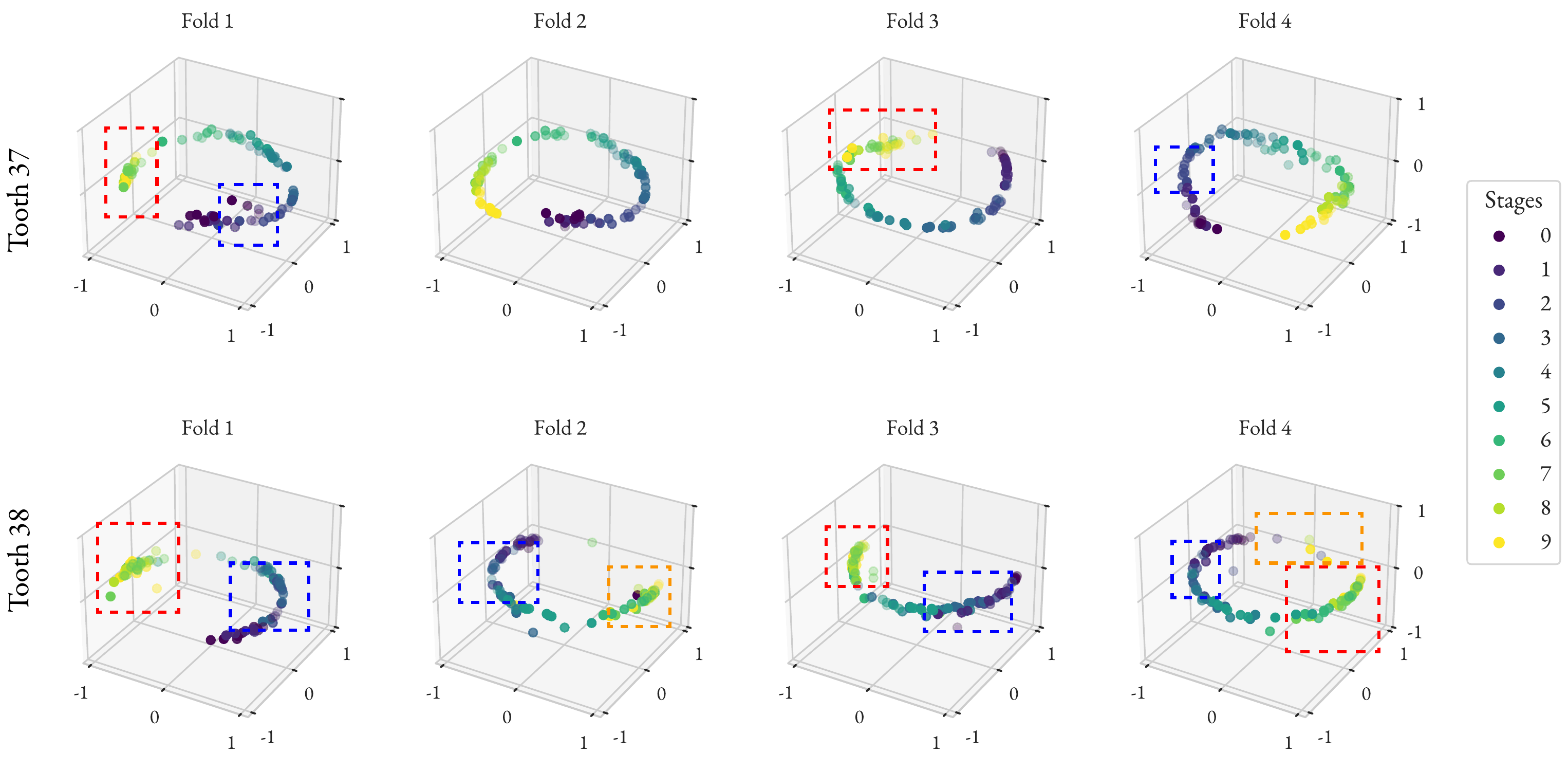}
    \caption{The PCA plots of latent space representations of the test set images, for both teeth from all folds. The embeddings are L2-normalized to mimic their contribution in the triplet loss during training. Dashed rectangles are utilised to highligh areas of high overlap of differing types. The blue rectangles indicate areas where there was high overlap between the embeddings of stages 0 to 5, a more common case for tooth 38. Red rectangles highlight highly-overlapping embeddings belonging to stages 6 to 10, which was encountered for both teeth. Orange rectangles mark areas of overlap between stages 0, 1, 8 and 9, which is only encountered with tooth 38. Generally, the embeddings of ordinally neighboring stages are grouped together, forming a smooth gradient from stage 0 to 9. However, for tooth 38, there is more overlap between stages, and discontinuities exist in the gradient, implying low inter-class variation.}
    \label{fig:38_bbox}
\end{figure}

In Fig. \ref{fig:38_bbox}, we can observe the latent space representations that are produced in each fold of the cross-validation process for the test set. 
The embeddings for tooth 37 consistently converge towards the desired structure, and generally these embeddings satisfy the variable-margin triplet condition. They form a smooth gradient from stage 0 to 9 round a 3-sphere, which is incentivised by L2-normalization during loss calculation in training. The embeddings which share a stage label are packed closer together, while maximizing the mean Euclidean distance between further classes, dictated by the margin. There are few to no discontinuities, the lack of which implies that the AE model could effectively optimize the pipeline by finding distinct features per stage in the input images. The main failure mode for tooth 37 latent space, i.e. the overlap between stages in the latent space, occurs at the later stages of 7 to 9, where the divisions become less clear. To a lesser degree, stages 2 to 3 also display this issue. However, it can be said that the embeddings generally conform well to the triplet condition well for this tooth. Since the latent space is the first bottleneck in the AE + ViT pipeline, inconsistencies in the embeddings have an effect on the rest of the process. More specifically, if there is a drastic overlap between two stages in the latent space, the reconstructions of these embeddings will also be visually similar, making the classification process more difficult. Indeed, the regions of overlap were also the stages which had more classification error (Fig. \ref{fig:violinsVit}). Accordingly, for tooth 38, we can see the overlaps are more drastic, and span multiple stages. Discontinuities in the gradient are also much more drastic. These observations indicate the AE model did indeed have difficulty in spreading out stages along the surface of a 3-sphere, failing to satisfy the triplet condition during training more often than for tooth 37. A failure mode unique to tooth 38 is the overlap between the embeddings of stages 0 and 9, which would result in visually similar reconstructions of these two highly distinct stages. These failures reveal that the AE model, across the folds, was unable to find common distinct features in the input images of tooth 38, hinting at a high intra-class variability and low inter-class variability for this dataset as the cause for the higher downstream error (Supplementary Fig. 6).

\begin{figure}[!h]
    \centering
    \includegraphics[width=.9\textwidth]{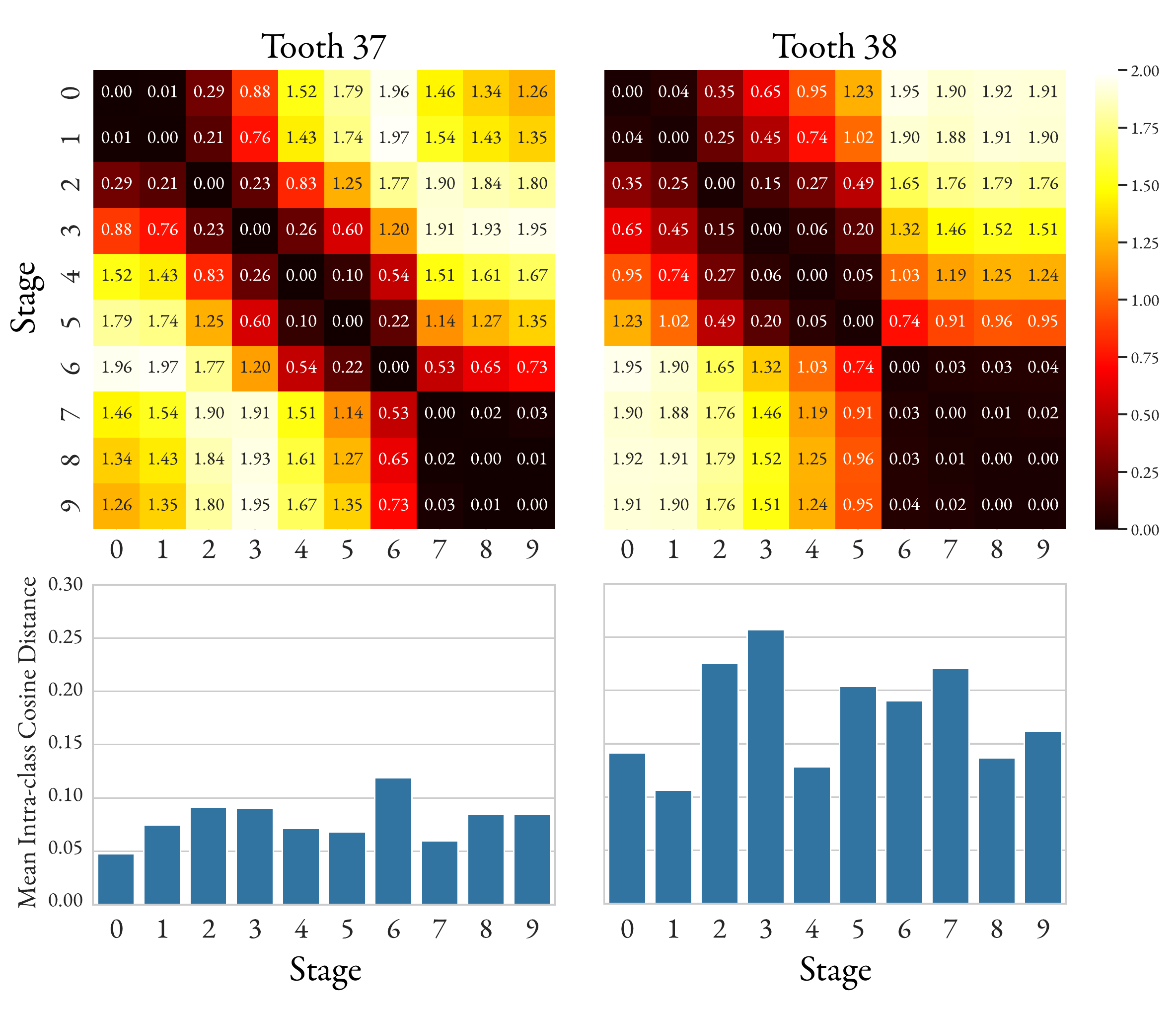}
    \caption{An analysis of the latent space grouping of the embeddings for both teeth. The embeddings from the test sets of all folds were used for the calculations. Top row shows the inter-class distances, defined as the pairwise cosine distance between the L2-normalized geometric centroids of each developmental stage, depicting how well the embeddings were separated from those of different stages. The separability of stages in the latent spaces for tooth 38 shows stages 0 to 5 and stages 6 to 9 are highly grouped together, an undesirable outcome which is less pronounced for tooth 37. Bottom row plots the mean intra-class pairwise cosine distances per stage, showing how well each stage is grouped together in the latent space. The lower values for tooth 37 than for tooth 38 translate to better grouping for tooth 37. }
    \label{fig:embd_comp}
\end{figure}

A quantitative analysis of the latent space for teeth 37 and 38, depicted in Fig. \ref{fig:embd_comp}, reveals significant differences in class separability and compactness. The inter-class distance heatmaps (Fig. 11, top row), which measure the separation between the mean embeddings of each stage, show a well-structured latent space for tooth 37. Distances are low between adjacent stages (e.g. stages 3 and 4 have a distance of 0.26) and progressively increase for non-adjacent stages, reflecting the ordinal nature of the data. In contrast, the heatmap for tooth 38 displays poor separation between several classes; for instance, stages 3, 4 and 5 show very low inter-class distances, indicating their embeddings are closely grouped together and difficult to distinguish. This poor grouping can be interpreted as a confidence measure in the model, beyond the performance metrics. Since the latent space lies closer to the input in our pipeline, the low inter-class variability in this representation is likely to be reflected in the reconstructions as well (Supplementary Fig. 3).

These observations are further clarified by mean intra-class cosine distances (Fig. 11, bottom row), which measure the compactness of samples within each stage. For tooth 37, the intra-class distances are consistently low, generally below 0.10, indicating that samples from the same stage were encoded in tight clusters, meeting the triplet demands imposed during training. Conversely, tooth 38 exhibits markedly higher intra-class separation across all stages, with notable peaks at stages 2, 3, and 7. The combination of low inter-class separation and high intra-class variability for tooth 38 provides a quantitative explanation for the challenges in its classification, as its latent space is demonstrably less structured and class-separable than that of tooth 37.

\subsection*{Synthesis of Findings and Contextualization in Literature}

A synthesis of the evidence from the proposed AE+ViT framework indicates that the poor performance on tooth 38, and the subsequent performance disparity between tooth 37 and tooth 38 is attributable not to a failure of the deep learning architecture, but to the intrinsic properties of the tooth 38 dataset. This conclusion addresses a critical challenge in medical AI, where standard interpretability methods can be misleading. The baseline ViT attention maps for tooth 38 exemplify this problem: they appeared plausible by focusing on relevant anatomical structures, yet they failed to explain the model’s poor performance or identify the diagnostically crucial features it was missing. This highlights the limitation of relying on a single, superficial mode of interpretability, which may be plausible but not faithful to the model's actual failure mode. For instance, Ong et al. (2024) reported a high accuracy of 0.905 for stagin teeth 36 and 37 into one of the eight original Demirjian stages using EfficientNet \cite{ong2024fully}. Since their Grad-CAM attention maps demonstrated high plausibility---with an attention shift over the stages corresponding with the human understanding of molar development---they failed to explain a slight drop in performance for stage G (= stages 7 and 8 in the current study), which may hinder adoption of the method in real-life, high-stakes use cases. By contrast, our proposed multi-faceted AE + ViT framework allows moving beyond a plausible explanation and uncovers the underlying issue. By applying our framework, we were able to meet the performance metrics reported by Matthijs et al. (2024) for tooth 38, increasing the accuracy from 0.462 to 0.543, and surpass them for tooth 37, achieving 0.815 accuracy and 0.252 MAE. Furthermore, the agreement of our model with the labels assigned by experts, quantified by $\kappa_w$, aligned well with the reported intra- and inter-rater agreements in the literature of studies using the Demirjian staging method. Maia et al. (2010)\cite{maia2010demirjian} reported an intra-observer $\kappa$ coefficient of 0.52 for second molars, highlighting the low agreement with the same rater. While not directly comparable to any automated method, including our pipeline, as the automation eliminates intra-rater disagreement completely, this highlights that the ratings can be quite variable with human labelers. As for studies reporting on inter-rater variability, the results still show a wide range of values. Ambarkove et al. (2014)\cite{ambarkova2014dental} reported an inter-rater $\kappa$ score of 0.7 for the second molars ---a score comparable to the $\kappa_w$ achieved by our framework for tooth 37 (0.79). Elshehawi et al. (2016) \cite{elshehawi2016dental} reported a $\kappa$ score of 0.77 in the inter-rater agreement in the staging of all third molars, and Merdietio et al. (2025)\cite{merdietio2025evaluation} reported a Gwet AC2 metric of 0.924 for tooth 37, indicating strong reliability. For specifically the third molars, a $\kappa$ score of 0.69 was reported by Boonpitaksathit et al. (2011)\cite{boonpitaksathit2011dental}. In light of this agreement metrics, while closer to the lower end of the distribution, our reported $\kappa_w$ of 0.79 for tooth 37 is within a plausible range. However, these results also highlight that the agreement of our model with the labelers, quantified by a $\kappa_w$ of 0.49 at most, is weak. This further contextualises the low performance on tooth 38. When the good labeler agreement of our framework on tooth 37 is also considered, we can more strongly conclude that the performance issue is data-centric. 

The 'diagnostic' capabilities of the framework provided converging lines of evidence. First, visual inspection of the AE outputs showed that tooth 38 reconstructions failed to conform to sharp stage prototypes and were considerably blurrier than those for tooth 37, qualitatively suggesting high morphological inconsistency in the source data (Fig. \ref{fig:ims_recons}). This observation was quantitatively substantiated by a direct analysis of the latent space metrics (Fig. \ref{fig:embd_comp}). The tooth 38 embeddings display not only markedly higher mean intra-class cosine distances, confirming a lack of compactness within stages, but also poor inter-class separation in the distance heatmaps, indicating a disorganized latent space where class boundaries are ill-defined. Finally, the AE+ViT attention maps for tooth 38 confirmed that even after preprocessing, the model learned to rely almost exclusively on the crown, suggesting the root regions in that dataset were too morphologically variable to provide reliable staging information. Therefore, the evidence points to the conclusion that the tooth 38 dataset is characterized by high intra-class variation and low inter-class separability. Supplementary Figs. 5 and 6 provide a visual comparison between selected original images of teeth 37 and 38, illustrating their variability. This claim is supported by the satisfactory classification performances seen with tooth 37, using DenseNet-201, ViT, and AE + ViT models, and further demonstrated by Matthijs et al. (2024)\cite{matthijs2024artificial}. This data-centric issue can be interpreted in three ways: (1) the specific sample collection fails to consistently represent each developmental stage, (2) the issue is rooted in the inherently greater morphological variability known for third molars, or (3) a combination of both factors. Ultimately, this investigation establishes the proposed framework not merely as a performance-enhancing tool, but as a crucial methodology for achieving diagnostic transparency. It demonstrates a necessary progression beyond plausible but superficial interpretability, enabling a robust, data-centric diagnosis of model failure in high-stakes forensic applications.

\section*{Limitations and Future Prospects}
This work has several limitations that warrant discussion. The modest size of the datasets ($\approx$ 400 images per tooth) is a known limiting factor in model generalization. Even with the gold standard selection process and train-time data augmentations, the sample size remains too low to meaningfully generalize to a larger population. For this, our framework needs to be evaluated, and potentially re-trained on a larger dataset. The singular origin of the data also may hinder generalization. All data was collected in the same institution, from a Belgian population, and as such is only representative of the patient profile of this institution. This limitation can be remedied by fine-tuning the model on a more diverse dataset, ideally collected at multiple institutions. In this context, distributed training strategies such as federated learning can be of use. Furthermore, our methodology has its own constraints. The smoothing effect of the autoencoder, while beneficial for the ViT, could obscure fine-grained diagnostic features. As such, the diagnostic reliability of the reconstruction images is diminished until the framework is trained with a larger dataset, which can rectify the quality of the reconstructions. Finally, while a consensus protocol was used for labeling, any model trained on a manual standard may be influenced by residual inter-observer variability in the reference labels. 
Although similar weighted $\kappa$ values have been reported for the reproducibility of molar staging on 3D imaging modalities compared to 2D approaches \cite{franco2020comparing,de2017forensic}, one would nonetheless expect, at least intuitively, a lower degree of inter-observer variability when assessing apical closure stages on dental (CB)CT than on OPG \cite{de2019magnetic}. This expectation arises because the morphological changes associated with apical closure are often subtle and may be obscured by artefacts inherent to OPG, whereas such artefacts are largely eliminated when visualized on sectional (CB)CT images. Analogous to how clavicle CT has progressively replaced conventional clavicle radiographs in the context of forensic age estimation, it would be worthwhile to investigate whether automated methods for dental assessment hold the potential to demonstrate a similar superiority of CBCT over OPG. However, it must be borne in mind that the explainability of these automated approaches should remain guaranteed. Finally, it should be emphasized that single-site age estimation ought to be avoided under all circumstances. A holistic approach to age estimation — whether manual or automated — should therefore always combine dental with skeletal predictors \cite{de2020dental}.

\section*{Conclusion}

In this work, we investigated the significant performance disparity observed in the automated dental staging of mandibular second (tooth 37) and third (tooth 38) molars. We introduced and evaluated a transparent deep learning framework, consisting of a convolutional autoencoder (AE) and a Vision Transformer (ViT), designed not only for classification but also to serve as a diagnostic tool for understanding model behaviour. Our proposed AE + ViT framework provided a multi-faceted diagnosis of the issue. Evidence from the AE in the form of blurry reconstructions in the root region, the disorganized latent space metrics and the crown-centric attention maps all converged on the conclusion that the dataset for tooth 38 is characterised by high intra-class and low inter-class variation. Therefore, we conclude that the poor staging performance on tooth 38 is not an architectural failure of the deep learning models but a data-centric problem rooted in the high intrinsic morphological variability of third molars. We backed this conclusion by demonstrating an identical AE + ViT pipeline on tooth 37, and showed that the classification performance, the attention maps and the latent space analysis all revealed a satisfactory classification in line with results from the current literature. This study establishes our proposed framework as an essential methodology for achieving diagnostic transparency and demonstrates a necessary progression beyond attention-based explanations, which were plausible but not informative.

In a practical workflow, this framework could function as a 'second opinion' tool for experts. For instance, when a low confidence prediction is made, our framework can provide the forensic odontologists with the reconstruction, the position of the image in the latent space, and attention maps. This information can then be further analysed, and reliability measures such as the nearest neighbors in the embedding space and attention similarity can be produced, which can aid the expert decision process quantitatively. This would provide data-driven evidence to confirm that the uncertainty of the model stems from tooth morphology, and not a systemic model error, thus adding a critical layer of context to their final assessment.

\section*{Data Availability}
The data used in this work has been gathered and processed via the local ethics approval obtained from the Ethical Commission Research UZ/KU Leuven with the approval number S62392. As such, the data cannot be openly shared. However, the data that will enable the reproduction of this work can be accessed upon a reasonable request, subject to evaluation by the ethical approval body, from author Jannick De Tobel.

\bibliography{myrefs}

\section*{Acknowledgements}

This work was funded by KU Leuven, under Internal Research Fund as C2-project number 3E180439. 
We would like to thank Katarzyna Koncewicz for her invaluable assistance in processing the clinical images and in creating the figures presented in this paper.



\section*{Additional information}

The authors declare no competing interests.

\end{document}